\documentclass[letterpaper]{article} %
\usepackage[preprint]{aaai2027-arxiv}
\usepackage{authblk}
\usepackage[hyphens]{url}  %
\usepackage{graphicx} %
\usepackage{natbib}  %
\usepackage{caption} %
\usepackage{algorithm}
\usepackage{algorithmic}
\usepackage{amsmath}
\usepackage{amsfonts}

\usepackage{newfloat}
\usepackage{listings}
\DeclareCaptionStyle{ruled}{labelfont=normalfont,labelsep=colon,strut=off} %
\floatstyle{ruled}
\newfloat{listing}{tb}{lst}{}
\floatname{listing}{Listing}

\usepackage{booktabs}
\usepackage{tabularx}

\title{GERF: A Token-Efficient Multimodal Retrieval Framework for Schema-Guided Scientific Information Extraction}

\author[1]{Yiyu Qian}
\author[2]{Xing Wu}
\affil[1]{Department of Computer Science, Cornell University\\
yq339@cornell.edu}
\affil[2]{School of Computer Engineering and Science, Shanghai University\\
xingwu@shu.edu.cn}
\affiliations{}

\begin{document}

\maketitle

\begin{abstract}
Scientific information extraction (IE) is a fundamental component of AI for Science, transforming experimental knowledge embedded in scientific literature into structured and traceable data for evidence synthesis, scientific database construction, and data-driven discovery. However, critical experimental details are often distributed across text, tables, figures, captions, and surrounding discussion, making them difficult to recover using conventional text-only extraction. Existing LLM-based and multimodal RAG pipelines typically rely on full-document context or coarse page-level retrieval, introducing substantial irrelevant evidence and high textual and visual token costs. We propose GERF, a Graph-based Evidence Retrieval Framework for token-efficient, schema-guided multimodal scientific IE. GERF represents each article as a heterogeneous evidence graph that connects schema fields, text chunks, figures, and figure-derived content through semantic and structural relations. Using schema fields as retrieval anchors, GERF combines direct semantic retrieval, Random Walk with Restart, and null-aware evidence expansion to construct compact, field-specific multimodal evidence packages. Visual interpretation is applied only to graph-selected figures, avoiding indiscriminate processing of the entire document. We evaluate GERF from two complementary perspectives: evidence retrieval quality and pipeline-level input-token consumption. On the expert-annotated SOFC-Exp corpus, GERF retrieves gold-supported evidence for 94.55\% of the evaluated field instances, providing broad field-level evidence coverage for downstream extraction. In the token comparison, GERF reduces average input-token usage by 73.40\% relative to full-context processing and uses fewer tokens than the evaluated page-level pipelines. These results establish GERF as a fine-grained and token-efficient evidence retrieval foundation for schema-guided scientific IE and broader AI-for-Science applications.
\end{abstract}

\section{Introduction}
Transforming scientific literature into structured and traceable data is an important foundation for AI for Science, enabling large-scale knowledge integration, automated hypothesis generation, and data-driven scientific discovery \cite{wang2023scientific,olivetti2020datadriven,dagdelen2024structured}. However, much of this information remains embedded in unstructured or semi-structured documents, making it difficult to aggregate, compare, and reuse at scale. Scientific information extraction (IE) aims to transform such literature into structured, machine-readable records, supporting evidence-based literature review, scientific database construction, knowledge graph generation, and data-driven discovery \cite{tshitoyan2019unsupervised,ammar2018construction,luan2018multitask}.

Scientific IE is challenging because research articles are long, densely written, and highly domain-specific. Evidence for a schema field may appear across multiple sections, and multi-value fields often require evidence from several non-adjacent passages. Important information is also communicated through figures, captions, tables, and their surrounding discussion. Flat chunk retrieval may identify an isolated relevant passage while missing complementary evidence elsewhere in the document. In contrast, full-context and page-level multimodal processing retain broader context but introduce substantial textual and visual token costs. Scientific IE therefore requires a retrieval mechanism that can recover distributed multimodal evidence while limiting the context supplied to downstream models.

To address this challenge, we propose GERF, a Graph-based Evidence Retrieval Framework for token-efficient, schema-guided multimodal scientific IE. GERF represents each paper as a heterogeneous evidence graph containing schema-field, text-chunk, figure, and figure-content nodes. Semantic and structural edges encode schema-to-text relevance, figure-to-text grounding, and local document relationships. Direct schema retrieval first identifies high-confidence evidence, after which Random Walk with Restart propagates relevance to related nodes. A null-aware refilling stage expands the search only for unresolved fields. GERF further limits visual-model usage to graph-selected figures and compacts the resulting multimodal evidence into a field-relevant representation for downstream extraction.

Our main contributions are as follows:

\begin{itemize}
    \item We introduce GERF, a heterogeneous graph-based retrieval framework that explicitly connects schema fields with textual and visual evidence, preserving cross-modal and document-level relationships that are lost in flat chunk or page-level retrieval.

    \item We develop a field-conditioned retrieval strategy that combines direct semantic retrieval, graph propagation, null-aware context expansion, and selective visual processing. This design targets distributed evidence, particularly for multi-value fields, while avoiding full-document multimodal input.

    \item We construct a schema-aligned evaluation benchmark from SOFC-Exp, a high-quality expert-annotated full-text corpus with fine-grained, source-grounded annotations of complex experimental entities, values, and relations. The resulting gold answer and source-evidence templates support field-level and value-level retrieval evaluation, together with pipeline-level comparisons of textual and visual token consumption.
\end{itemize}

\paragraph{Paper Organization}
The remainder of this paper is organized as follows. We first review
related work on multimodal document retrieval, graph-based retrieval,
and scientific information extraction. We then present GERF, beginning
with the construction of the heterogeneous evidence graph, followed by
schema-conditioned evidence propagation, selective multimodal evidence
assembly, and null-aware evidence expansion for unresolved fields.
Next, we describe the SOFC-Exp-based evaluation design and assess GERF
from two complementary perspectives: evidence retrieval quality and
pipeline-level token efficiency. Finally, we summarize the main findings
and discuss directions for future work.

\section{Related Work}

Multimodal retrieval has progressed from shared image--text representations toward unified and fine-grained retrieval architectures. UniIR learns an instruction-guided retriever across multiple multimodal tasks \cite{wei2024uniir}, while PreFLMR uses late interaction between query tokens, visual features, and document representations \cite{lin2024preflmr}. MuRAG further incorporates multimodal retrieval into retrieval-augmented generation by retrieving images and text for knowledge-intensive question answering \cite{chen2022murag}. Although effective for cross-modal matching, these approaches are not specifically designed for schema-guided scientific IE, where one field may require complementary evidence distributed across several textual and visual sources \cite{shamsabadi-etal-2024-large,khalighinejad-etal-2025-matvix}.

Recent work retrieves directly from visually rich documents. ColPali represents rendered pages with vision-language models and applies late interaction for page retrieval \cite{faysse2025colpali}. VisRAG similarly retrieves pages in their visual form to preserve layout and graphical information \cite{yu2025visrag}, while M3DocRAG extends page retrieval to multi-page and multi-document question answering \cite{cho2024m3docrag}. MMDocIR evaluates retrieval over both pages and layout elements, including text, tables, figures, and equations \cite{dong2025mmdocir}. These methods preserve information that may be lost during text-only parsing, but page-level retrieval can introduce substantial irrelevant context. Moreover, independently ranked pages or layout elements do not explicitly preserve the relationships among a figure, its caption, and the paragraphs required to interpret it.

GERF specializes multimodal retrieval for schema-guided scientific IE. It represents schema fields, text chunks, figures, and figure-derived content as a heterogeneous evidence graph, with captions and document structure used to establish semantic and structural connections. Schema fields serve as retrieval anchors, while graph propagation recovers complementary evidence connected through figure references and surrounding context \cite{tong2006fast,gutierrez2024hipporag}. By assembling field-specific evidence and processing only graph-selected figures, GERF preserves cross-modal relationships without passing complete pages or the full document to the downstream model.

\section{Method}

GERF consists of three principal stages. First, GERF constructs
a heterogeneous graph over schema fields, text chunks, and figures.
Second, it identifies high-confidence textual anchors for each schema
field and propagates their relevance through the graph. Finally, the
direct and propagated evidence is assembled into a compact multimodal
evidence subgraph for downstream extraction.

\subsection{Heterogeneous Evidence Graph Construction}

Given a scientific document $D$ and an extraction schema
$\mathcal{S}=\{f_1,\ldots,f_M\}$, the objective is to retrieve, for
each schema field $f_m$, a compact set of multimodal evidence that is
sufficient for downstream structured extraction.

We formulate this problem as field-conditioned retrieval over a
heterogeneous evidence graph
\begin{equation}
    \mathcal{G}=(\mathcal{V},\mathcal{E}),
\end{equation}
where $\mathcal{V}$ contains schema fields and document evidence units,
and $\mathcal{E}$ captures their semantic and structural relationships.
For each field $f$, GERF aims to retrieve a compact evidence subgraph
\begin{equation}
    \mathcal{H}_f \subseteq \mathcal{G},
\end{equation}
such that $\mathcal{H}_f$ preserves the complementary textual and visual
evidence relevant to $f$ while excluding unrelated document content.

\subsection{Node Construction}

GERF represents each scientific document as a heterogeneous evidence
graph that separates extraction targets from textual and visual
evidence. The graph contains four types of nodes
\begin{equation}
    \mathcal{V}
    =
    \mathcal{V}_{S}
    \cup
    \mathcal{V}_{T}
    \cup
    \mathcal{V}_{F}
    \cup
    \mathcal{V}_{FC},
\end{equation}
where $\mathcal{V}_{S}$ denotes schema-field nodes,
$\mathcal{V}_{T}$ text-chunk nodes, $\mathcal{V}_{F}$ figure nodes,
and $\mathcal{V}_{FC}$ figure-content nodes. The graph is initially
constructed from schema fields and document elements, and is later
augmented with figure-content nodes for figures selected as potentially
relevant to the extraction schema.

\paragraph{Schema-field nodes.}
Each leaf field in the extraction schema is represented as a node
$v_f \in \mathcal{V}_{S}$. A schema-field node contains the hierarchical
field path and a natural-language description of the information to be
extracted. Field-specific lexical cues, such as expected units, domain
terms, and common value expressions, may also be associated with the
node. We represent a schema-field node as
\begin{equation}
    v_f = \left(f, d_f, \mathcal{K}_f\right),
\end{equation}
where $f$ is the field identifier, $d_f$ is its description, and
$\mathcal{K}_f$ denotes optional field-specific lexical cues. These nodes
define the information needs that guide subsequent evidence retrieval.

\paragraph{Text-chunk nodes.}
The body text of the document is segmented into locally coherent,
non-overlapping chunks. Each chunk is represented as a node
$v_i \in \mathcal{V}_{T}$
\begin{equation}
    v_i = \left(x_i, p_i, s_i, o_i\right),
\end{equation}
where $x_i$ is the textual content, $p_i$ and $s_i$ denote its page and
section, and $o_i$ records its position in document order. Preserving
document provenance enables GERF to model local discourse continuity
and structural proximity during edge construction. Non-evidential
elements, such as page furniture, section headings, references, and
back matter, are excluded before text nodes are created.

\paragraph{Figure nodes.}
Each captioned scientific figure is represented as a node
$v_g \in \mathcal{V}_{F}$
\begin{equation}
    v_g = \left(I_g, c_g, \ell_g, p_g\right),
\end{equation}
where $I_g$ is the figure image, $c_g$ is its caption, $\ell_g$ is its
document identifier, and $p_g$ is its page location. The caption is
retained as an attribute of the figure node because it provides an
initial textual description of the visual content. However, the complete
meaning of a figure may also depend on the passages that reference and
interpret it.

\paragraph{Figure-content nodes.}
A figure-content node $v_h \in \mathcal{V}_{FC}$ represents the visual
information derived from a selected figure using a vision-language
model. Unlike a figure node, which preserves the original image and its
document metadata, a figure-content node stores a structured semantic
interpretation of the visible content, including observations,
candidate field values, relevant panels or regions, and uncertainty
information. We denote it as
\begin{equation}
    v_h = \left(z_h, g_h\right),
\end{equation}
where $z_h$ is the structured visual interpretation and $g_h$ identifies
its source figure. Figure-content nodes are created only for figures
retained by the retrieval stage, rather than for every image in the
document. A single figure-content node may provide evidence for multiple
schema fields, allowing the visual interpretation of the same figure to
be reused.

Together, these node types distinguish four complementary roles:
schema-field nodes specify \emph{what} information is required,
text-chunk nodes preserve explicit and explanatory statements, figure
nodes retain the original visual evidence and document context, and
figure-content nodes expose the semantic information contained within
selected figures.

\subsection{Edge Construction}

GERF constructs typed edges to capture direct field relevance, local
discourse structure, and cross-modal grounding. The complete edge set is
\begin{equation}
    \mathcal{E}
    =
    \mathcal{E}_{ST}
    \cup
    \mathcal{E}_{TT}
    \cup
    \mathcal{E}_{TF}
    \cup
    \mathcal{E}_{FC},
\end{equation}
where $\mathcal{E}_{ST}$ denotes schema--text edges,
$\mathcal{E}_{TT}$ text--text edges, $\mathcal{E}_{TF}$ text--figure
edges, and $\mathcal{E}_{FC}$ figure--content edges. The first three
edge types constitute the initial retrieval graph. Figure--content
edges are added only after relevant figures have been selected and
visually interpreted.

\paragraph{Schema--Text Edges.}
Schema--text edges encode the direct relevance of a text chunk to a
target schema field. For each field $f$, we construct an expanded query
$q_f$ from its hierarchical field name, natural-language description,
expected units, and optional field-specific terminology.

We independently encode the query and each candidate text chunk using
SciBERT \cite{beltagy2019scibert}, and compute their relevance using a
ColBERT-style late-interaction function \cite{khattab2020colbert}. Let
\begin{equation}
    \mathbf{Q}_f
    =
    \left[
        \mathbf{q}_{f,1},\ldots,\mathbf{q}_{f,m}
    \right]
\end{equation}
and
\begin{equation}
    \mathbf{X}_i
    =
    \left[
        \mathbf{x}_{i,1},\ldots,\mathbf{x}_{i,n}
    \right]
\end{equation}
denote the normalized token representations of the field query and text
chunk $x_i$, respectively. Their semantic relevance is computed using
MaxSim
\begin{equation}
    s_{\mathrm{MaxSim}}(f,i)
    =
    \frac{1}{m}
    \sum_{a=1}^{m}
    \max_{1\leq b\leq n}
    \mathbf{q}_{f,a}^{\top}\mathbf{x}_{i,b}.
    \label{eq:maxsim}
\end{equation}
For each query token, MaxSim identifies the most similar token in the
candidate chunk and averages these maximum similarities across the
query. This fine-grained interaction is useful for scientific
documents, where a long passage may support a field through only a few
technical terms, units, or value expressions.

Dense semantic matching may nevertheless underemphasize exact
scientific expressions. We therefore compute an auxiliary lexical score
$s_{\mathrm{lex}}(f,i)$ based on exact matches with field-name terms,
expected units, salient description terms, and optional field-specific
vocabulary. The schema--text edge weight is defined as
\begin{equation}
    w_{ST}(f,i)
    =
    s_{\mathrm{MaxSim}}(f,i)
    +
    \lambda_{\mathrm{lex}}
    s_{\mathrm{lex}}(f,i),
    \label{eq:schema_text}
\end{equation}
where $\lambda_{\mathrm{lex}}$ is a fixed hyperparameter that balances
exact lexical evidence against semantic MaxSim relevance. It is not a
learned parameter. For each field, only the highest-scoring candidate
chunks are retained
\begin{equation}
    \mathcal{E}_{ST}(f)
    =
    \operatorname{TopK}_{x_i\in\mathcal{V}_T}
    w_{ST}(f,i).
\end{equation}
These directed edges identify the most likely locations of direct
field evidence and later determine the starting points of
schema-conditioned propagation.

\paragraph{Text--Text Edges.}
Directly relevant chunks may omit qualifiers, experimental conditions,
or explanations expressed in nearby passages. GERF therefore connects
text chunks within a local document window. For two chunks $x_i$ and
$x_j$, the edge weight is
\begin{equation}
    w_{TT}(i,j)
    =
    \exp\left(
        -\lambda_T \Delta(i,j)
    \right)
    \rho(i,j),
    \label{eq:text_text}
\end{equation}
where $\Delta(i,j)=|o_i-o_j|$ is their distance in document order and
$\lambda_T$ controls the positional decay. The section-aware factor is
\begin{equation}
    \rho(i,j)
    =
    \begin{cases}
        1,
        & s_i=s_j,\\
        \gamma,
        & s_i\neq s_j,
    \end{cases}
    \qquad 0<\gamma<1,
\end{equation}
where $s_i$ and $s_j$ denote the corresponding document sections.
The factor $\gamma$ reduces propagation across section boundaries,
where positional adjacency is less likely to indicate continuous
evidence. Text--text edges are undirected and are retained only within
a limited neighborhood, preserving local discourse continuity without
densely connecting unrelated parts of the document.

\paragraph{Text--Figure Edges.}
Text--figure edges ground visual evidence in the passages that reference
or interpret it. For a figure $g$, we first identify the chunks that
explicitly mention its identifier
\begin{equation}
    \mathcal{A}_g
    =
    \left\{
        x_i\in\mathcal{V}_T
        \mid
        x_i \text{ explicitly references } g
    \right\}.
\end{equation}
These chunks are treated as high-confidence figure anchors and are
connected to the corresponding figure with unit weight
\begin{equation}
    w_{TF}(i,g)=1,
    \qquad
    x_i\in\mathcal{A}_g.
\end{equation}
An explicit reference alone, however, does not capture the full
discussion of a figure. Subsequent passages may describe its observed
patterns, explain individual panels, or interpret its scientific
meaning without repeating the figure identifier. To retrieve such
passages, we construct a figure-grounding query
\begin{equation}
    q_g
    =
    \operatorname{concat}
    \left(
        c_g,
        a_g
    \right),
    \label{eq:figure_query}
\end{equation}
where $c_g$ is the figure caption and $a_g$ is the first chunk that
explicitly references the figure. The anchor is kept fixed rather than
being updated by subsequently retrieved passages. When no explicit
reference is available, the caption alone is used as the grounding
query.

For each non-anchor chunk $x_i$, we compute semantic relevance using the
same SciBERT-based, ColBERT-style MaxSim function introduced in
Equation~\ref{eq:maxsim}. The text--figure edge weight is
\begin{equation}
    w_{TF}(i,g)
    =
    s_{\mathrm{MaxSim}}(q_g,x_i)
    \cdot
    d(i,g)
    \cdot
    \pi(i,g),
    \label{eq:text_figure}
\end{equation}
where $d(i,g)$ incorporates structural proximity to the nearest explicit
figure anchor, and $\pi(i,g)$ suppresses ambiguous associations with
competing figures.

We use a plateau-shaped distance decay
\begin{equation}
    d(i,g)
    =
    \begin{cases}
        1,
        & \Delta(i,g)\leq r,\\[5pt]
        \displaystyle
        \frac{1}
        {1+\frac{\Delta(i,g)-r}{h}},
        & \Delta(i,g)>r,
    \end{cases}
    \label{eq:plateau_decay}
\end{equation}
where $\Delta(i,g)$ is the chunk distance to the nearest anchor of
figure $g$, $r$ defines a local no-decay region, and $h$ controls the
decay beyond that region. The plateau reflects the observation that
several consecutive chunks following a figure reference may belong to
the same figure discussion.

The ambiguity factor $\pi(i,g)\in(0,1]$ decreases the edge weight when
the candidate chunk explicitly references another figure or lies closer
to another figure's anchor. These cases are penalized rather than
strictly removed, allowing shared or comparative figure discussions to
remain reachable. Non-anchor edges with insufficient relevance are
discarded. The resulting text--figure relations connect each figure not
only to its explicit citation, but also to the nearby passages needed
to interpret its visual content.

\paragraph{Figure--Content Edges.}
After schema-conditioned propagation selects potentially relevant
figures, each unique figure is interpreted once using a
vision-language model. The resulting figure-content node $v_h$ is
connected to its source figure $v_g$ by
\begin{equation}
    (v_g,v_h)\in\mathcal{E}_{FC},
    \qquad
    w_{FC}(g,h)=1.
\end{equation}
Unlike the preceding edge types, this weight does not represent semantic
similarity. It is a deterministic provenance relation that links the
original image to its structured visual interpretation. A shared
figure-content node may support multiple schema fields, avoiding
repeated visual processing of the same figure. Figure--content edges
augment the retrieved evidence after propagation and do not participate
in the initial random walk.

\subsection{Schema-Conditioned Evidence Propagation}

The schema--text edges identify passages that are directly relevant to
each extraction field, but they do not necessarily recover all evidence
needed to interpret those passages. Supporting conditions may occur in
neighboring chunks, while relevant figures may be connected only through
the text that references or explains them. GERF therefore propagates
field relevance over the evidence graph using Random Walk with Restart
(RWR) \cite{tong2006fast}.

\paragraph{Direct evidence anchors.}
For each schema field $f$, we first select the highest-scoring
schema--text connections
\begin{equation}
    \mathcal{A}_f
    =
    \operatorname{TopK}_{x_i\in\mathcal{V}_T}
    w_{ST}(f,i),
    \label{eq:direct_anchors}
\end{equation}
where $\mathcal{A}_f$ is the set of direct textual anchors for field
$f$. These anchors represent the passages most likely to explicitly
contain or describe the target value.

Rather than initiating propagation from the schema node itself, GERF
constructs a weighted restart distribution over the direct anchors
\begin{equation}
    p_f(i)
    =
    \begin{cases}
        \displaystyle
        \frac{w_{ST}(f,i)}
        {\sum_{x_j\in\mathcal{A}_f}w_{ST}(f,j)},
        & x_i\in\mathcal{A}_f,\\[10pt]
        0,
        & \text{otherwise}.
    \end{cases}
    \label{eq:restart_distribution}
\end{equation}
This design separates direct semantic matching from structural
expansion. The schema first identifies where the target information is
most likely to appear, and graph propagation then retrieves evidence
that supports or contextualizes those locations. Seeding from text
anchors also prevents the walk from propagating through shared chunks
into unrelated schema fields.

\paragraph{Type-aware transition matrix.}
Propagation is performed over the text and figure nodes using
text--text and text--figure edges. Because these edge types represent
different relations, their weights are rescaled before normalization
\begin{equation}
    \widetilde{w}^{(f)}_{ij}
    =
    \begin{cases}
        \lambda_{TT} w_{TT}(i,j),
        & (v_i,v_j)\in\mathcal{E}_{TT},\\[3pt]
        \lambda_{TF}^{(f)} w_{TF}(i,j),
        & (v_i,v_j)\in\mathcal{E}_{TF},\\[3pt]
        0,
        & \text{otherwise},
    \end{cases}
    \label{eq:type_scaled_edges}
\end{equation}
where $\lambda_{TT}$ controls propagation through local textual context,
and $\lambda_{TF}^{(f)}$ controls propagation between text and figures.
The latter may depend on the target field: visual-result fields and
fields that can be supported by schematics receive stronger
text--figure propagation, whereas ordinary text-dominant fields assign
less weight to visual paths.

The scaled edge weights are row-normalized to form the transition
matrix
\begin{equation}
    W^{(f)}_{ij}
    =
    \frac{\widetilde{w}^{(f)}_{ij}}
    {\sum_k \widetilde{w}^{(f)}_{ik}}.
    \label{eq:transition_matrix}
\end{equation}
Text--text and text--figure relations are treated as bidirectional
during propagation. This allows relevance to move from a direct textual
anchor to a referenced figure and from that figure to other passages
that explain it.

\paragraph{Random Walk with Restart.}
Let $\mathbf{r}^{(t)}_f$ denote the relevance distribution over text and
figure nodes at iteration $t$. GERF updates the distribution according
to
\begin{equation}
    \mathbf{r}^{(t+1)}_f
    =
    \alpha\mathbf{p}_f
    +
    (1-\alpha)
    \left(\mathbf{W}^{(f)}\right)^{\top}
    \mathbf{r}^{(t)}_f,
    \label{eq:rwr}
\end{equation}
where $\mathbf{p}_f$ is the anchor-based restart distribution and
$\alpha\in(0,1)$ is the restart probability. The restart term repeatedly
returns probability mass to the direct evidence, limiting drift toward
nodes that are structurally connected but semantically unrelated to the
field. The propagation term allows relevance to reach nearby context and
cross-modal evidence through the graph.

Iteration continues until convergence, yielding the stationary
distribution
\begin{equation}
    \mathbf{r}_f
    =
    \alpha\mathbf{p}_f
    +
    (1-\alpha)
    \left(\mathbf{W}^{(f)}\right)^{\top}
    \mathbf{r}_f.
    \label{eq:rwr_stationary}
\end{equation}
The stationary score \(r_f(v)\) reflects both the restart mass assigned
to direct field evidence and the accumulated probability of reaching
\(v\) through field-conditioned graph transitions. Consequently, nodes
connected to strong anchors through short, high-weight paths generally
receive higher relevance scores.

Importantly, GERF does not replace direct evidence with the RWR ranking.
The anchor set $\mathcal{A}_f$ is always retained as the primary
evidence. Immediate neighboring chunks are preserved to avoid losing
conditions or qualifiers split across chunk boundaries, while
high-scoring non-anchor text nodes provide additional contextual
evidence. The propagated figure scores are used as structural support
signals for selecting figures connected to the field's direct or
contextual text evidence. Thus, RWR functions as a controlled evidence
expansion mechanism rather than as an independent extractor.

\subsection{Multimodal Evidence Assembly}

For each schema field $f$, propagation produces direct textual anchors
$\mathcal{A}_f$, supporting context chunks $\mathcal{C}_f$, and candidate
figures $\mathcal{F}_f$. Direct anchors are retained as primary evidence,
while contextual chunks and figures supply complementary information
connected through the evidence graph.

\paragraph{Selective Visual Interpretation.}
GERF applies visual interpretation only to graph-selected figures. For
each candidate figure $g$, the VLM receives its image $I_g$, caption
$c_g$, linked text $\mathcal{L}_g$, and associated field descriptions
$\mathcal{S}_g$:

\begin{equation}
z_g
=
\Phi
\left(
I_g,
c_g,
\mathcal{L}_g,
\mathcal{S}_g
\right),
\label{eq:visual_interpretation}
\end{equation}
where $\Phi$ denotes the VLM and $z_g$ contains the resulting visual
interpretation and field-specific observations. Each unique figure is
processed once, and its output is stored as a shared figure-content node
that can support multiple schema fields.

\paragraph{Field-Specific Visual Verification.}
Because structural relevance does not guarantee visible support, GERF
verifies every field--figure association
\begin{equation}
\chi_f(g)
=
\begin{cases}
1, & \text{if figure } g \text{ visually supports field } f,\\
0, & \text{otherwise}.
\end{cases}
\end{equation}
Only positively verified figures are retained
\begin{equation}
\mathcal{F}^{+}_f
=
\left\{
g\in\mathcal{F}_f
\mid
\chi_f(g)=1
\right\},
\label{eq:verified_figures}
\end{equation}
with their corresponding figure-content nodes defined as
\begin{equation}
\mathcal{Z}_f
=
\left\{
z_g
\mid
g\in\mathcal{F}^{+}_f
\right\}.
\end{equation}
This step removes figures that are connected to relevant discussion but
do not contain usable visual evidence for the target field.

\paragraph{Compact Evidence Graph.}
The final field-specific evidence representation is

\begin{equation}
\mathcal{H}_f
=
\mathcal{A}_f
\cup
\mathcal{C}_f
\cup
\mathcal{F}^{+}_f
\cup
\mathcal{Z}_f,
\label{eq:compact_evidence}
\end{equation}
combining direct text, supporting context, verified figures, and their
visual interpretations. Shared nodes are reused across fields, while
retrieval scores and unsupported visual associations are excluded. The
downstream extractor receives only the field definition and
$\mathcal{H}_f$, rather than the complete document.

\paragraph{Null-Aware RWR Expansion.}

The initial evidence graph may yield a null output when relevant evidence
falls beyond the first-pass retrieval cutoff. GERF treats such fields as
unresolved rather than assuming that the requested information is absent.
For each unresolved field, a second field-conditioned RWR pass retains
the original direct anchors but applies a larger evidence budget to
retrieve additional text and figure nodes. Formally, let
$\mathcal{H}^{(0)}_f$ and $\hat{y}^{(0)}_f$ denote the initial evidence
graph and extraction result. The refilled evidence graph is
\begin{equation}
\mathcal{H}^{(1)}_f
=
\begin{cases}
\begin{aligned}
&\mathcal{H}^{(0)}_f \cup {}\\
&\operatorname{Top}_{K_{\mathrm{refill}}}
\left(
\mathcal{V}\setminus\mathcal{H}^{(0)}_f;
r^{(1)}_f
\right),
\end{aligned}
&
\hat{y}^{(0)}_f=\texttt{null},
\\[4pt]
\mathcal{H}^{(0)}_f,
&
\text{otherwise}.
\end{cases}
\label{eq:null_refill}
\end{equation}
where $r^{(1)}_f$ is the second-pass RWR score and
$K_{\mathrm{refill}}$ is the expanded retrieval budget. Newly selected
figures undergo the same field-specific visual verification before being
merged with the existing evidence. Only unresolved fields are
re-extracted, improving evidence coverage without increasing the context
or visual-processing cost for fields that have already been filled.

\section{Experiments}
\label{sec:experiments}

\subsection{Experimental Setup}
\label{sec:experimental_setup}

The experiments evaluate GERF from two complementary perspectives: retrieval quality and pipeline-level token efficiency. The retrieval-quality experiment evaluates whether GERF can locate evidence containing the required gold values, while the token-efficiency experiment compares the multimodal input cost of different document-processing pipelines.

\begin{table}[t]
\centering
\caption{Main experimental configuration.}
\label{tab:experimental_configuration}
\begin{tabularx}{\columnwidth}{lX}
\toprule
\textbf{Item} & \textbf{Configuration} \\
\midrule
Hardware & Apple M4, 16 GB unified memory \\
Operating system & macOS 26.3.1 \\
Environment & Python 3.14.1; Python 3.10 for ColPali \\
Acceleration & CPU and PyTorch MPS with float16 \\
Multimodal model & Qwen3-VL-Plus through DashScope \\
Page retrievers & VisRAG-Ret and ColPali v1.2 \\
\bottomrule
\end{tabularx}
\end{table}

\subsection{Dataset and Evaluation Design}
\label{sec:dataset_evaluation}

We use the SOFC-Exp corpus \cite{friedrich-etal-2020-sofc}, an expert-annotated benchmark for scientific information extraction in the materials-science domain. The corpus contains 45 open-access articles describing solid oxide fuel cell experiments. Domain experts annotated experiment expressions, materials, devices, conditions, measurements, and their semantic roles using token-level BIO labels.

To support schema-guided retrieval evaluation, we deterministically transform the original BIO and slot annotations into a structured gold answer template. Each schema field is associated with its annotated gold values and source evidence, including the supporting sentence, span offsets, entity type, and slot label. This template provides a standardized reference for determining whether the evidence retrieved for a field contains the required gold information. After excluding metadata fields and values without grounded textual evidence, the resulting benchmark contains 936 evaluated field instances and 3,805 required gold values. No training or fine-tuning is performed, and all retrieval methods are evaluated in a zero-shot setting against the same gold answer template.

\subsubsection{Retrieval-Quality Ablation}

We compare three GERF configurations:

\begin{enumerate}
    \item \textbf{Direct Retrieval}: retrieves the five highest-relevant text chunks for each schema field without graph-based contextual expansion.
    \item \textbf{RWR}: uses the directly retrieved chunks as seeds and retrieves additional contextual evidence through Random Walk with Restart.
    \item \textbf{RWR + Null Refill}: applies an expanded RWR evidence
    budget only to schema fields that remain null after the initial
    extraction pass.
\end{enumerate}

For each schema field, the gold answer template records not only the correct values but also all source evidence chunks that support each value. The evidence returned by each pipeline is then compared with the annotated source evidence in the gold template. A gold value is considered successfully retrieved if at least one of its associated source evidence chunks appears in the retrieved evidence set.

Formally, let $G_f$ denote the set of gold values for field $f$, $S(g)$ the set of annotated source evidence chunks supporting gold value $g$, and $R_f$ the chunks retrieved by a pipeline for field $f$. The retrieval indicator for a gold value is defined as
\begin{equation}
h(g,R_f)=
\mathbb{I}\left[R_f \cap S(g) \neq \emptyset\right].
\end{equation}
Based on this definition, \textit{Value Recall} measures the proportion of individual gold values for which at least one supporting chunk is successfully retrieved:
\begin{equation}
\mathrm{ValueRecall}
=
\frac{
\sum_{f}\sum_{g\in G_f}h(g,R_f)
}{
\sum_f |G_f|
}.
\end{equation}
\textit{Any-Value Hit} measures the proportion of evaluated fields for which the retrieved evidence supports at least one gold value:
\begin{equation}
\mathrm{AnyValueHit}
=
\frac{1}{|\mathcal{F}|}
\sum_{f\in\mathcal{F}}
\mathbb{I}
\left[
\sum_{g\in G_f}h(g,R_f)>0
\right].
\end{equation}
In contrast, \textit{All-Values Hit} applies a stricter field-level criterion and requires the retrieved evidence to cover every gold value associated with the field:
\begin{equation}
\mathrm{AllValuesHit}
=
\frac{1}{|\mathcal{F}|}
\sum_{f\in\mathcal{F}}
\mathbb{I}
\left[
\sum_{g\in G_f}h(g,R_f)=|G_f|
\right].
\end{equation}
We additionally report the average number of retrieved chunks per field to quantify the evidence budget required by each pipeline. This metric distinguishes improvements obtained through more effective evidence retrieval from those achieved simply by returning substantially more document context.

\subsubsection{Pipeline-Level Token Comparison}

We compare GERF against three multimodal document-processing baselines:

\begin{itemize}
\item \textbf{Full Context} processes every figure and supplies the complete parsed document to the schema-filling prompt.
\item \textbf{VisRAG-style} retrieves relevant PDF pages, converts each selected page into a textual description, and uses the generated descriptions to construct a downstream schema-filling prompt.
\item \textbf{M3DocRAG-style} uses ColPali to retrieve relevant pages and supplies the selected page images directly to a multimodal reader, without intermediate page-to-text conversion.
\item \textbf{GERF} uses graph-based retrieval to construct a compact evidence package and applies the VLM only to schema-relevant figures selected by the evidence graph.
\end{itemize}

All pipelines are evaluated SOFC-Exp corpus using the same extraction schema and paper-level accounting protocol. For the two page-level baselines, each schema field retrieves its top two pages. The resulting pages are unioned and deduplicated within each paper and rendered as 144-dpi RGB images. VisRAG-style retrieval performs page-level VLM analysis followed by one schema-filling prompt, whereas M3DocRAG-style retrieval constructs one multimodal reader prompt containing the selected page images.

For each paper, the input cost is decomposed as
\begin{equation}
T_{\mathrm{input}}
=
T_{\mathrm{text}}
+
T_{\mathrm{image}},
\end{equation}
where $T_{\mathrm{text}}$ includes instructions, schema content, retrieved textual evidence, and intermediate page descriptions, while $T_{\mathrm{image}}$ represents the visual tokens contributed by figures or PDF pages.

Only input tokens required to construct the multimodal extraction context are counted. Output generation and local operations, including PDF processing, embedding computation, similarity scoring, graph construction, and RWR propagation, are excluded. Token reduction relative to full-context processing is calculated as
\begin{equation}
R_{\mathrm{token}}
=
1-\frac{T_{\mathrm{method}}}{T_{\mathrm{full}}}.
\end{equation}

\subsection{Retrieval-Quality Results}
\label{sec:retrieval_results}

\begin{table}[t]
\centering
\caption{Retrieval-quality ablation on SOFC-Exp.}
\label{tab:retrieval_ablation}
\resizebox{\columnwidth}{!}{
\begin{tabular}{lcccc}
\toprule
\textbf{Method} &
\textbf{Value Recall} &
\textbf{All Hit} &
\textbf{Any Hit} &
\textbf{Chunks/Field} \\
\midrule
Direct Retrieval
& 62.47\% & 49.57\% & 89.96\% & 5.00 \\
RWR
& 69.41\% & 55.66\% & 93.06\% & 13.77 \\
RWR + Null Refill
& \textbf{74.30\%} & \textbf{59.94\%}
& \textbf{94.55\%} & 21.64 \\
\bottomrule
\end{tabular}}
\end{table}

Table~\ref{tab:retrieval_ablation} shows that direct retrieval achieves 89.96\% Any-Value Hit with only five chunks per field, but its Value Recall and All-Values Hit remain at 62.47\% and 49.57\%. This gap demonstrates the necessity of context expansion for multi-value fields, whose supporting evidence is often distributed across different document regions.

RWR improves Value Recall to 69.41\%, while null-aware refilling further raises it to 74.30\%. The complete GERF configuration retrieves 2,827 of 3,805 gold values, finds gold-supported evidence for 885 of 936 evaluated fields, and retrieves all annotated values for 561 fields. Compared with direct retrieval, it recovers approximately 450 additional gold values and completely covers 97 additional fields. Most notably, its 94.55\% Any-Value Hit rate provides a highly reliable evidence foundation for downstream extraction without requiring full-document input.

This improvement increases the average evidence budget from 5.00 to 21.64 chunks per field, reflecting a coverage-budget trade-off. However, the expanded evidence remains schema-guided and graph-selected, allowing GERF to recover distributed evidence without blindly passing the complete document to the downstream model.

\subsection{Token-Efficiency Results}
\label{sec:token_results}

\begin{table}[t]
\centering
\caption{Average input-token composition per paper on SOFC-Exp.}
\label{tab:token_comparison}
\resizebox{\columnwidth}{!}{
\begin{tabular}{lrrrr}
\toprule
\textbf{Pipeline} &
\textbf{Prompt/Text} &
\textbf{Image} &
\textbf{Total} &
\textbf{Reduction} \\
\midrule
Full Context
& 51,049.74 & 36,449.26 & 87,499.00 & -- \\
VisRAG-style
& 23,287.41 & 18,158.25 & 41,445.66 & 52.63\% \\
M3DocRAG-style
& 12,119.30 & 13,742.45 & 25,861.75 & 70.44\% \\
GERF
& \textbf{12,943.63} & \textbf{10,332.00}
& \textbf{23,275.63} & \textbf{73.40\%} \\
\bottomrule
\end{tabular}}
\end{table}

As shown in Table~\ref{tab:token_comparison}, full-context processing requires 87,499 input tokens per paper because it includes both the complete textual document and all visual content. VisRAG-style retrieval reduces this cost by 52.63\%, but still consumes 23,287.41 textual tokens because its page descriptions are included again in a downstream schema-filling prompt.

M3DocRAG-style retrieval avoids this intermediate page-to-text stage and reduces the total cost to 25,861.75 tokens. GERF achieves the lowest overall cost at 23,275.63 tokens per paper. Although its textual cost is slightly higher than that of M3DocRAG-style retrieval, its visual cost is 24.82\% lower because GERF processes only graph-selected, schema-relevant figures rather than complete retrieved pages.

Importantly, GERF does not achieve its efficiency by shifting cost from one modality to another. Relative to full-context processing, it simultaneously reduces textual and visual input by 74.65\% and 71.65\%, respectively, requiring only 26.60\% of the original input-token budget. This balanced reduction produces a 73.40\% decrease in total input tokens, while GERF also requires 43.84\% fewer tokens than VisRAG-style retrieval and approximately 10.00\% fewer than M3DocRAG-style retrieval.

\section{Conclusion}

This paper presented GERF, a graph-based framework for token-efficient, schema-guided multimodal scientific information extraction. By connecting schema fields, text chunks, figures, and figure-derived content within a heterogeneous evidence graph, GERF combines direct retrieval, graph propagation, null-aware expansion, and selective visual processing to recover distributed evidence without exhaustive document processing. On SOFC-Exp, GERF retrieves 74.30\% of annotated gold values and identifies gold-supported evidence for 94.55\% of the evaluated fields, while using only 26.60\% of the full-context input budget. It reduces textual, visual, and total input tokens by 74.65\%, 71.65\%, and 73.40\%, respectively, and remains more token-efficient than the evaluated page-level pipelines. These results show that fine-grained graph retrieval can preserve cross-modal evidence relationships while providing broad evidence coverage at substantially lower cost. GERF therefore offers a practical and traceable retrieval foundation for transforming scientific literature into structured knowledge. Future work will extend GERF to additional scientific domains and integrate it with end-to-end extraction and verification systems. We will also explore its application to large-scale scientific database construction, automated literature synthesis, and knowledge graph population. These extensions will help establish GERF as a more general and scalable foundation for transforming multimodal scientific literature into structured, traceable, and verifiable knowledge.

\bibliography{aaai2027}

\end{document}